\documentclass[runningheads]{llncs}

\usepackage{eccv}

\usepackage{eccvabbrv}

\usepackage{multirow} 
\usepackage{graphicx}
\usepackage{amssymb}% http://ctan.org/pkg/amssymb
\usepackage{pifont}% http://ctan.org/pkg/pifont
\usepackage{marvosym}
\usepackage{graphicx}
\usepackage{booktabs}

\usepackage[accsupp]{axessibility}  % Improves PDF readability for those with disabilities.

\usepackage{hyperref}

\usepackage{orcidlink}

\begin{document}

% ---------------------------------------------------------------
% TODO REVIEW: Replace with your title
\title{Harmonizing knowledge Transfer in Neural Network with Unified Distillation}

% TODO REVIEW: If the paper title is too long for the running head, you can set
% an abbreviated paper title here. If not, comment out.
\titlerunning{Harmonizing knowledge Transfer in Neural
Network with Unified Distillation}

% TODO FINAL: Replace with your author list. 
% Include the authors' OCRID for the camera-ready version, if at all possible.
\author{Yaomin Huang\inst{1}\orcidlink{0000-0002-8195-4978} \and
Zaomin Yan\inst{1} \and
Chaomin Shen\inst{1}\orcidlink{0000-0001-9389-6472}
Faming Fang\inst{1} \and \\
Guixu Zhang\inst{1}\textsuperscript{, \Letter}
}

% TODO FINAL: Replace with an abbreviated list of authors.
\authorrunning{Huang et al.}
% First names are abbreviated in the running head.
% If there are more than two authors, 'et al.' is used.

% TODO FINAL: Replace with your institution list.
\institute{School of Computer Science, East China Normal University \\
\email{\{ymhuang, ZaoMingYan\}@stu.ecnu.edu.cn} \\
\{cmshen, fmfang, gxzhang\}@cs.ecnu.edu.cn
\footnote[0]{\textsuperscript{\Letter}~Corresponding author.}}

\maketitle
\begin{abstract}
Knowledge distillation (KD), known for its ability to transfer knowledge from a cumbersome network (teacher) to a lightweight one (student) without altering the architecture, has been garnering increasing attention.
Two primary categories emerge within KD methods: feature-based, focusing on intermediate layers' features, and logits-based, targeting the final layer's logits.
This paper introduces a novel perspective by leveraging diverse knowledge sources within a unified KD framework.
Specifically, we aggregate features from intermediate layers into a comprehensive representation, effectively gathering semantic information from different stages and scales.
Subsequently, we predict the distribution parameters from this representation. 
These steps transform knowledge from the intermediate layers into corresponding distributive forms, thereby allowing for knowledge distillation through a unified distribution constraint at different stages of the network, ensuring the comprehensiveness and coherence of knowledge transfer.
Numerous experiments were conducted to validate the effectiveness of the proposed method.
\end{abstract}    
\section{Introduction}
\label{sec:intro}
\begin{figure}[t]
\centering
\includegraphics[width=\columnwidth]{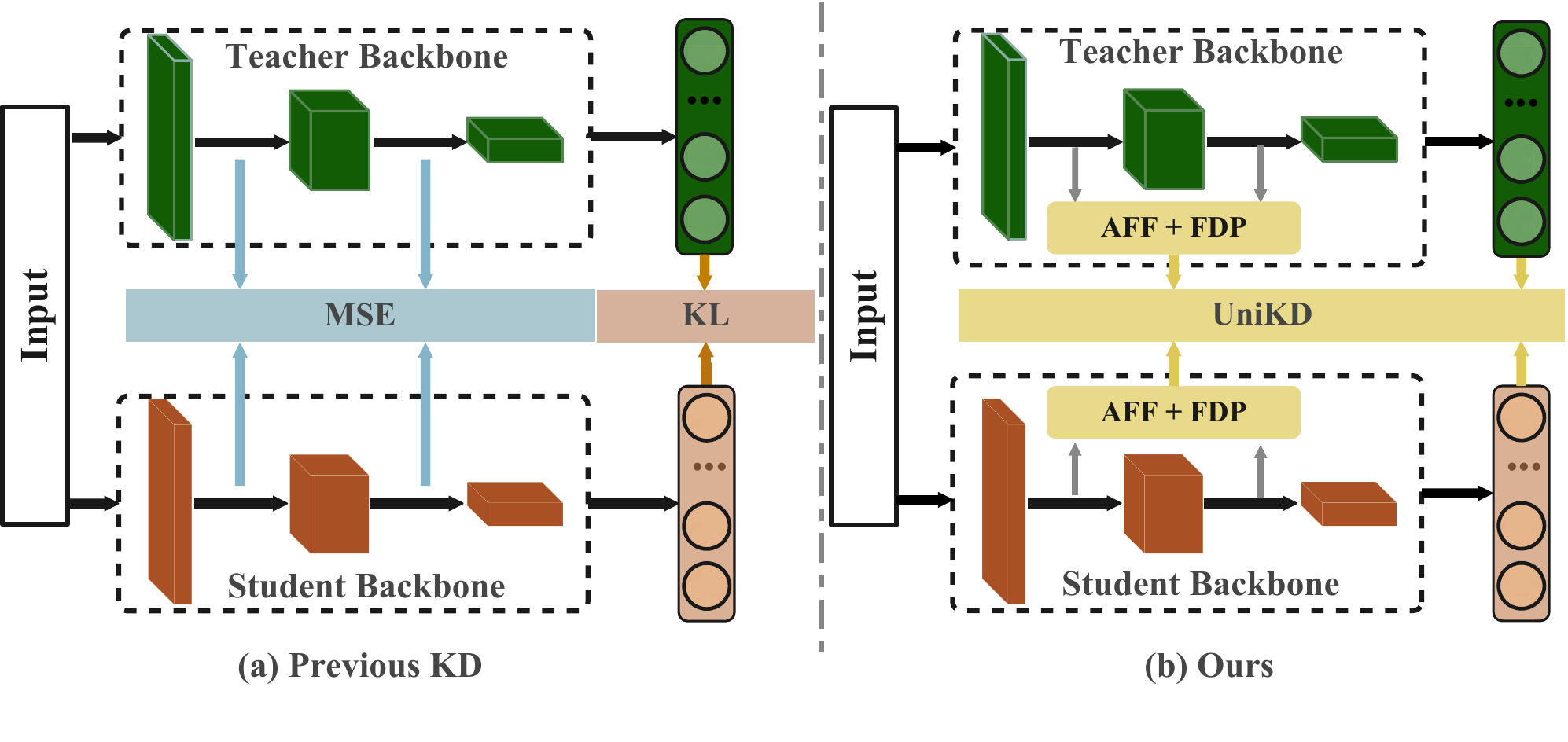}
\caption{Comparison of Methodologies in feature-based knowledge distillation (MSE), logits-based knowledge distillation (KL), and the proposed UniKD.
UniKD achieves unified knowledge distillation across different network layers, considering information at various levels while circumventing the incoherence from direct combinations.}
\label{Introduction}
\vspace{-0.1cm}
\end{figure}
Despite the considerable efficacy of large-scale models~\cite{raffel2020exploring, brown2020language, chowdhery2022palm, touvron2023llama} in a broad spectrum of tasks, their substantial size and computational demands present significant challenges.
Knowledge Distillation, a widely employed network compression technique, continues to attract substantial attention.
The primary objective of knowledge distillation is to facilitate knowledge transfer from a sophisticated, high-capacity teacher network to a more streamlined and efficient student network.
Mainstream knowledge distillation methods are primarily classified into two categories: 1) logits-based~\cite{cho2019efficacy, zhao2022decoupled, yang2023knowledgewsld, jin2023multi} and 2) feature-based~\cite{romero2014fitnets, lin2022knowledge, li2023rethinking}.
Logits-based distillation focuses on transferring the teacher's output classification probability to the student network.
This methodology focuses on capturing the class probability distribution, which serves as distribution-level constraints.
Conversely, feature-based approaches seek to align the individual feature values or activations at each spatial location between the teacher and student networks.
This process can be viewed as imposing pixel-level constraints.

A critical review of current knowledge distillation methods reveals that these approaches typically center on either a singular type of knowledge~\cite{hinton2015distilling, romero2014fitnets, Zhang_2018_CVPR, zhou2021wsl, zhao2022decoupled, zagoruyko2016paying, tian2019contrastive, chen2021distilling} or a direct hybridization of two knowledge types~\cite{DBLP:journals/tip/SongCYS22, DBLP:conf/aaai/HeoLY019a}, overlooking the inconsistencies in knowledge across different layers.
This paper aims to facilitate knowledge transfer from the teacher to the student network across various layers using a unified constraint.
We will explain the proposed method using three step-by-step problems.
First,\textbf{ why conduct knowledge distillation at different layers?}
On the one hand, considering feature capture at different backbone stages is crucial. Shallow layers encode robust local features, bolstering the network's noise and interference resistance, while deep layers capture abstract semantic information, enhancing adaptability to novel data. On the other hand, the final layer's logits, which directly relate to the task's decision-making, contain rich information. Thus, the integration of different stages features and the utilization of logits' knowledge are necessary.
Second, \textbf{why unify different types of knowledge distillation methods?}
Continuous-valued features are compared in the continuous space using Mean Square Error (MSE), while logits consider the shape of the overall distribution via Kullback-Leibler (KL) divergence rather than just the difference in individual values. These two distillation methods have fundamentally different optimization objectives, which lead to varying effects on the gradients and update rules of the network parameters. 
Attempting to have the student network receive different types of supervision from the teacher in a single distillation process can result in unclear optimization objectives, making it difficult to achieve the optimal solution.
Third,\textbf{ why unify different knowledge into logits}?
Features contain a large amount of information, and each element includes information on the input data. However, this information can be redundant or only indirectly related to the final task. 
Ensuring strict pixel-level feature alignment between the teacher and student networks can impose overly rigid constraints and result in sub-optimal solutions~\cite{shu2021channel}. 
In contrast, each dimension of logits directly corresponds to a specific category, and its value reflects the network's prediction information for that category. Logits carry a higher information density and are directly related to the final decision process.

Building upon the previous analysis,  we introduce the \textbf{Uni}fied \textbf{K}nowledge \textbf{D}istillation (UniKD), which ensures harmonized knowledge transfer across different layers, thus facilitating an integrated and coherent transfer from the teacher to the student network.
Specifically, the Adaptive Features Fusion (AFF) module is employed to derive an integrated representation from intermediate layers.
AFF can retain critical information across different scales while simplifying the computation process.
Following this, we explore the knowledge distribution at intermediate layers by utilizing the Feature Distribution Prediction (FDP) module to estimate distribution parameters. This way, knowledge distillation within the intermediate layers can be conducted by implementing distribution-level constraints.
Upon completing these steps, UniKD enables a uniform and coherent knowledge distillation process through the network's various layers.
In summary, the main contributions of this work include:
\begin{itemize}
    \item We comprehensively analyze the challenges in current knowledge distillation methods and introduce UniKD. This framework ensures uniform knowledge distillation across various network layers, thereby ensuring the integrity and coherence of the knowledge transfer.
    \item The AFF module aggregates features from intermediate layers, which retains multi-scale information while avoiding unnecessary information transfer and further utilizes the FDP to harmonize the knowledge across different layers.
    \item The superiority of our method is demonstrated by extensive comparative experiments with teacher-student pairs across various datasets for different tasks.
\end{itemize}
\section{Related Work}
\label{sec:related}
\begin{figure*}[t]
\centering
\includegraphics[width=0.99\textwidth]{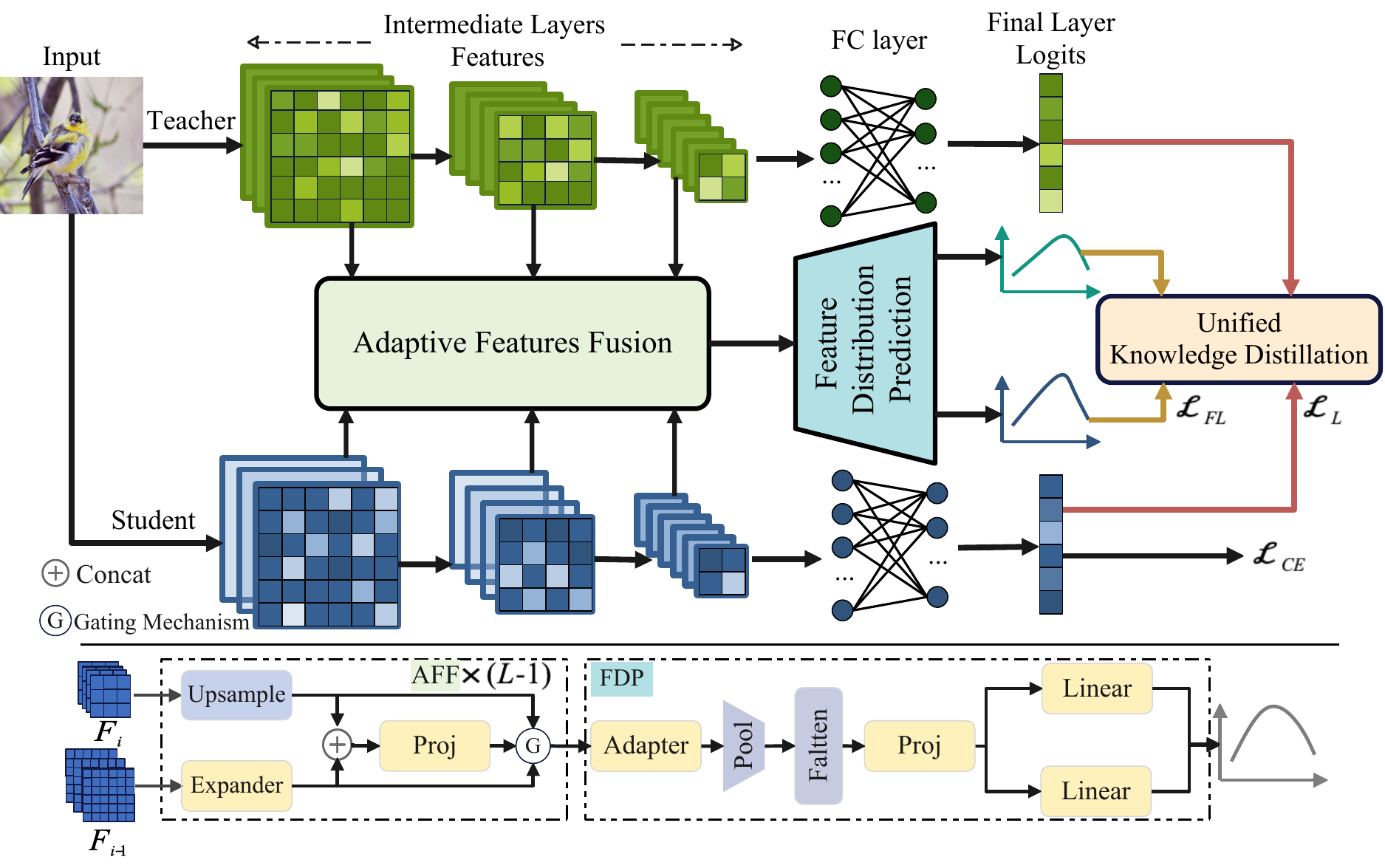}
\caption{The overall framework of the proposed UniKD. Our approach primarily comprises two modules. Initially, the features from the intermediate layers are aggregated using the stacked Adaptive Features Fusion (AFF) modules, followed by the derivation of the corresponding feature distribution through the Feature Distribution Prediction module (FDP). Consequently, our \textbf{Uni}fied \textbf{K}nowledge \textbf{D}istillation (UniKD) facilitates knowledge distillation in both the intermediate layer's features distribution and the final layer's probability distribution under a unified constraint.}
\label{framework}
\vspace{-0.1cm}
\end{figure*}
\subsection{Logits-based Knowledge Distillation}
Knowledge Distillation, first introduced in~\cite{hinton2015distilling}, aims to distill knowledge from a large, cumbersome model into a smaller, faster one without significant generalization performance loss.
It encourages the student model to mimic the teacher's outputs by minimizing the divergence between the teacher's and student's predictions.
Following this, a significant amount of work has focused on the research of logits-based knowledge distillation.
DML~\cite{Zhang_2018_CVPR} introduces a mutual-learning paradigm that enables an ensemble of students to learn collaboratively and teach each other throughout the training process.
WSL~\cite{zhou2021wsl} explores the reasons behind the functionality of soft labels by comparing the bias-variance of direct training without distillation and training with distillation.
DKD~\cite{zhao2022decoupled} and USKD~\cite{yang2023knowledgewsld} decouple the predicted class distributed to further extract information from the logits.
MLKD~\cite{jin2023multi} aligns logits not only at the instance level but also at the batch and class levels.
~\cite{Zhang_2019_ICCV} achieves self-distillation by converting the final logits and the features from intermediate layers into logits using a fully connected layer.

\subsection{Feature-based Knowledge Distillation}
Unlike distillation on logits, feature-based methods were first introduced in FitNets~\cite{romero2014fitnets}, which transfers knowledge from intermediate features.
Inspired by this, various methods have been proposed to match the features.
AT~\cite{zagoruyko2016paying} focuses on transferring the attention of feature maps to the students.
OFD~\cite{heo2019comprehensive} operates by collaboratively considering the adaptive layer, distillation feature, and distance function.
CRD~\cite{tian2019contrastive} aims to maximize the lower bound of mutual information between the teacher and student by utilizing the approach of contrastive learning.
KR~\cite{chen2021distilling} enhances distillation performance through the review of existing knowledge.
Additionally, some methods~\cite{park2019relational, tung2019similarity} concentrate on distilling the correlation to convey the teacher's knowledge.

Previous KD approaches focused on only one of these types of knowledge (features or logits). Some studies, despite incorporating two knowledge types~\cite{DBLP:journals/tip/SongCYS22, hsu2022closer}, need to address the distinctions between features and logits.
~\cite{wang2023crosskd} converts features into corresponding logits using the head, disregarding the differences in knowledge across different layers.
This paper proposes UniKD, a solution that enables consistent knowledge transfer across different layers through unified knowledge.

\section{Unified Knowledge Distillation}
\label{sec:method}
This section begins with a review of existing distillation methods, analyzing their limitations. Subsequently, we present our method, detailing the specific modules it comprises. Figure~\ref{framework} provides a comprehensive overview of this framework.
\subsection{Preliminaries}
\paragraph{Logits-based KD} was first proposed in~\cite{hinton2015distilling}, where the student learns the logits knowledge from the teacher through KL divergence.
In contrast to direct supervised learning through ground truth, logits-based KD using soft labels can elicit more `dark knowledge'~\cite{hinton2015distilling}, thereby enhancing the performance of the student network without altering its architecture.
The logits distillation losses are represented as follows:
\begin{align}
\label{eq_logits}
{\cal L}_{L}=K L(p^{t}||p^{s})=\sum_{j=1}^{C}p_{j}^{t}(\tau) \log (\frac{p_{j}^{t}(\tau)}{p_{j}^{s}(\tau)}),  
\end{align}
$P_j$ denotes the class probability derived from the logits $z \in \mathbb{R}^{C}$ after undergoing Softmax. $\tau$ is the temperature scaling hyper-parameter which enables the production of different probability distributions:
\begin{align}
    P_{j}(\tau)=\frac{e^{z_{j}/\tau}}{\sum_{c=1}^{C}e^{z_{c}/\tau}}.
\end{align}
\paragraph{Feature-based KD} transfers the knowledge of intermediate layers features from the teacher to the student. Generally, we can formulate such distillation methods as:
\begin{align}
{\cal L}_{F}=
\frac{1}{2}\sum_{i=1}^{n}||\mathcal{F}_i^{t}-w(\mathcal{F}_i^s)||^2,
\end{align}
where $\mathcal{F}_i^{t}$ and $\mathcal{F}_i^{s}$ are the corresponding features of the teacher and student at $i$-th layer. $w(\cdot)$ is a mapping function that aligns the dimensions of features from the student to the teacher.

From the above distillation process, it can be seen that logits-based knowledge distillation focuses on constraining the overall knowledge distribution. In contrast, feature-based knowledge distillation imposes pixel-level constraints on the student network. These two methods have distinct optimization objectives during the distillation process, and directly integrating them can lead to sub-optimal solutions.
In this study, we aim to achieve comprehensive knowledge transfer by conducting knowledge distillation on the intermediate layer features and the final layer logits.
To address this issue, we introduce UniKD, a meticulously designed framework that adjusts the intermediate layer features for a thorough and consistent knowledge transfer, using the same constraints employed for the final layer's logits.

\subsection{Unified Knowledge Transfer}
\label{3.2}
In this paper, we convert the intermediate layers features at the pixel level into a distribution form of type logits, facilitating the distillation of knowledge at the distribution level from the intermediate layers. 
This primarily considers the inherent differences in representation between the teacher and student networks due to the gap in model size, making it difficult for the student network to understand excessive pixel-level constraints. In contrast, logits provide a higher-level abstract representation that captures the overall characteristics of the input data rather than just the local pixel-level details. Therefore, by aligning the logits distributions, the student network can better learn the knowledge from the teacher network, enhancing its performance.

\subsubsection {Adaptive Features Fusion Module}
Directly transforming features from each layer to their corresponding distributions is computationally expensive and challenging. Therefore, we first fuse features from different layers to obtain a feature that incorporates multi-scale information. This approach ensures that the resultant feature representation captures both fine-grained details from lower layers and high-level semantics from upper layers. By integrating multi-scale information, the model can effectively process diverse features with reduced computational overhead.
We propose the Adaptive Features Fusion (AFF) module, which aggregates feature information from different layers. 
We denote the output of each intermediate layer's last residual blocks as ${\{\mathcal{F}_1,\mathcal{F}_2,\dots,\mathcal{F}_L\}}$.
For $\mathcal{F}_{i-1}$, its channels are increased through an expander and then concatenated with the result after upsampling of $\mathcal {F}_{i}$.
Subsequently, both information is fused through a gate mechanism, and the integrated intermediate representation $\mathcal{F}_R$ are output through an top-down method.
Formally,
\begin{equation} \label{eqaff}
  \begin{split}
  \mathcal{F}_{R} &= \mathbf{g} \odot \mathbf{E}(\mathcal{F}_{i-1})+ (1-\mathbf{g}) \odot \mathbf{Up}(\mathcal{F}_{i}),\\
  \mathbf{g} &= f(\mathbf{E}(\mathcal{F}_{i-1}) \oplus \mathbf{Up}(\mathcal{F}_{i})),
  \end{split}
\end{equation}
where $i={2, 3, \dots, L}$, $\mathbf{E}$ represents a $1 \times 1$ convolution to adapt to the number of channels at the $(i-1)$-th layer, $\mathbf{Up}$ denotes an upsampling operation to match the resolution of the $(i-1)$-th layer, $f$ stands for a gating function implemented through a $3 \times 3$ convolution, $\odot$ indicates point-wise multiplication, and $\oplus$ represents the concatenation operator. By adaptively determining the importance of adjacent layer features through $g$, this approach ensures that the resultant $\mathcal{F}_{R}$ retains the relatively significant components across different layers while eliminating redundant information.

\subsubsection{Feature Distribution Prediction Module}
Feature-based Knowledge distillation inherently steers the student's intermediate feature towards convergence with the teacher's, suggesting a shared underlying distribution for student and teacher features. 
In this work, we align the features of the teacher and student networks at the intermediate layers to a multivariate Gaussian distribution. The Gaussian assumption allows us to model the distribution of features with fewer parameters, capturing the essential characteristics of the data while discarding redundant information. Thus, this approach not only facilitates computation by leveraging the properties of Gaussian distributions, which simplify mathematical operations and statistical analysis, but has also been empirically proven to be effective in improving the performance of the student network.

For the multivariate Gaussian distributions corresponding to the features output by the student network, $q_\phi=\mathcal{N}\left(\mu_s, \Sigma_s\right)$, and those output by the teacher network, $q_\theta=\mathcal{N}\left(\mu_t, \Sigma_t\right)$, the KL divergence between them can be determined with an explicit analytical solution. $\mu_s$ and $\mu_t$ represent the mean vectors of the distributions corresponding to student and teacher features, respectively, while $\Sigma_s$ and $\Sigma_t$ denote the corresponding covariance matrices. The analytical solution as follows:
\begin{align}
\resizebox{.9\hsize}{!}{$
    \mathcal{L}_{FL}\left(q_\phi \| p_\theta\right)=\frac{1}{2}\left(\operatorname{tr}\left(\Sigma_t^{-1} \Sigma_s\right)+\left(\mu_t-\mu_s\right)^{\top} \Sigma_t^{-1}\left(\mu_t-\mu_s\right)-k+\ln \left(\frac{\left|\Sigma_t\right|}{\left|\Sigma_s\right|}\right)\right)$},
    \label{kl_as}
\end{align}
where $tr(\cdot)$ denotes the trace of a matrix, $|\cdot|$ represents the determinant of the matrix, and $k$ is the dimensionality of the features. A detailed proof of the process for obtaining the explicit analytical solutions is provided in the Supplementary Materials.

Eq. \ref{kl_as} involves the computation of the inverse matrix, the determinant, as well as the trace of the matrix, which is not only time-consuming but also numerically unstable, particularly in high-dimensional spaces. Here, we simplify the corresponding covariance matrices to diagonal matrices, specifically $\Sigma_s = \operatorname{diag}\left(\sigma_{s 1}^2, \sigma_{s 2}^2, \ldots, \sigma_{s k}^2\right)$ and $\Sigma_t = \operatorname{diag}\left(\sigma_{t 1}^2, \sigma_{t 2}^2, \ldots, \sigma_{t k}^2\right)$. Consequently, Eq. \ref{kl_as} can be simplified as:
\begin{align}
    \mathcal{L}_{FL}\left(q_\phi \| p_\theta\right)=\frac{1}{2} \sum_{i=1}^k\left(\frac{\sigma_{s i}^2}{\sigma_{t i}^2}+\frac{\left(\mu_{t i}-\mu_{s i}\right)^2}{\sigma_{t i}^2}-1+\ln \left(\frac{\sigma_{t i}^2}{\sigma_{s i}^2}\right)\right),
    \label{fkl_simple}
\end{align}
where $\sigma_{si}^2$ and $\sigma_{ti}^2$ represent the variance of the student and teacher features in the $i$th dimension, respectively, while $\mu_{si}$ and $\mu_{ti}$ denote the corresponding means.

Based on the aforementioned analysis, we only need to predict the corresponding mean values $\mu_s$, $\mu_t$ and variances $\sigma_{s}^2$, $\sigma_{t}^2$ of the student and teacher network features.
Regarding the $\mathcal{F}_{R}$ obtained through the Attention Feature Fusion (AFF) method, we deploy a Feature Distribution Prediction (FDP) module with shared parameters to predict the parameters of the multivariate Gaussian distributions corresponding to the fused features of both the student $\mathcal{F}_{R}^s$ and the teacher $\mathcal{F}_{R}^t$. Specifically, for $\mathcal{F}_{R}^s$ and $\mathcal{F}_{R}^t$, we apply pooling and flattening operations, followed by a projection to align the dimensions with the task label, and then predict the mean values $\mu_s$, $\mu_t$ and variances $\sigma_{s}^2$, $\sigma_{t}^2$. This process implements the distillation of feature knowledge as formulated in Eq.~\ref{fkl_simple}.

\subsection{Overall Objective Loss}
We employ the AFF method to merge features from various layers, subsequently converting them into their respective distributions using the FDP. This method effectively standardizes the process of distilling knowledge from both the intermediate layers and the final layer logits. Our approach ensures a consistent and efficient knowledge transfer by utilizing KL divergence, as detailed in Eq.~\ref{fkl_simple} and \ref{eq_logits} respectively. This not only enhances the learning capacity of the student model by leveraging the rich feature information from the teacher model but also simplifies the overall distillation process.
\begin{table*}[tp]
\centering
\caption{Results on the CIFAR-100 validation. Teachers and students are in homogeneous architectures. Top-1 accuracy is adopted as the evaluation metric and original performance is reported respectively. NKD$^*$ denotes NKD~\cite{sun2024logit} with DKD~\cite{zhao2022decoupled}.}\label{cifar_homo}
\begin{tabular}{cc|cccccc} 
\toprule
\multirow{2}{*}{Model}    & Teacher~ & ResNet56 & ResNet110 & ResNet32$\times$4 & WRN-40-2 & WRN-40-2 & VGG13  \\
                          & Student  & ResNet20 & ResNet32  & ResNet8 $\times$4 & WRN-16-2 & WRN-40-1 & VGG8   \\ 
\midrule
\multirow{2}{*}{Baseline} & Teacher                            & 72.34    & 74.31     & 79.42  & 75.61    & 75.61    & 74.64  \\
                          & Student                            & 69.06    & 71.14     & 72.50  & 73.26    & 72.98    & 70.36  \\ 
\midrule
\multirow{4}{*}{Feature}  & FitNet~\cite{romero2014fitnets}    & 69.21    & 71.06     & 73.50  & 73.58    & 72.24    & 71.02  \\
                          & CRD~\cite{tian2019contrastive}     & 71.16    & 73.48     & 75.51  & 75.48    & 74.14    & 73.94  \\
                          & AT~\cite{zagoruyko2016paying}      & 70.55    & 72.31     & 73.44  & 74.08    & 72.77    & 71.43  \\
                          & KR~\cite{chen2021distilling}       & 71.89    & 73.89     & 75.63  & 76.12    & 75.09    & 74.84  \\ 
\midrule
\multirow{3}{*}{Logits}   & KD~\cite{hinton2015distilling}     & 70.66    & 73.08     & 73.33  & 74.92    & 73.54    & 71.98  \\
                          
                          & DKD~\cite{zhao2022decoupled}       & 71.97    & 74.11     & 76.32  & 76.24    & 74.81    & 74.68  \\ 
                          & NKD$^*$~\cite{sun2024logit}        & 72.32    & 74.29     & 77.01  & 76.39    & 74.89    & 74.81  \\
\midrule
\multicolumn{2}{c|}{Ours}                               & \textbf{73.01}    & \textbf{75.36}     & \textbf{78.18}   & \textbf{77.49}    & \textbf{75.35}    & \textbf{75.76}  \\
\bottomrule
\end{tabular}
\end{table*}
In summary, 
by integrating Eq.~\ref{eq_logits} and Eq.~\ref{fkl_simple},
the total training objective for the student model is:
\begin{equation}
\label{loss_all}
    \mathcal{L}_{total} =
    \mathcal{L}_{CE} + \alpha \mathcal{L}_{FL} + \beta \mathcal{L}_{L},
\end{equation}
where $\mathcal{L}_{CE}$ is the standard task training loss for the student,
$\alpha, \beta$ is corresponding weights.
From $\mathcal{L}_{total}$, it can be observed that, apart from the standard supervision loss of the student network, all other losses are implemented through KL divergence. Consequently, this approach permits knowledge distillation within a comparatively unified paradigm.
\section{Experiments}
\label{sec:exp}
\begin{table*}[tp]
\centering
\caption{Results on the CIFAR-100 validation. Teachers and students are in heterogeneous architectures. Top-1 accuracy is adopted as the evaluation metric and original performance is reported respectively. NKD$^*$ denotes NKD~\cite{sun2024logit} with DKD~\cite{zhao2022decoupled}.}\label{cifar_hete}
\scalebox{0.9}{\begin{tabular}{cc|cccccc} 
\toprule
\multirow{2}{*}{Model}    & Teacher~ & ResNet32$\times$4 & WRN-40-2       & VGG13        & ResNet50     & ResNet32$\times$4   \\
                          & Student  & ShuffleNet-V1     & ShuffleNet-V1  & MobileNet-V2 & MobileNet-V2 & ShuffleNet-V2       \\ 
\midrule
\multirow{2}{*}{Baseline} & Teacher                            & 79.42    & 75.61     & 74.64  & 79.34    & 79.42      \\
                          & Student                            & 70.50    & 70.50     & 64.60  & 64.60    & 71.82      \\ 
\midrule
\multirow{4}{*}{Feature}  & FitNet~\cite{romero2014fitnets}    & 73.54    & 73.73     & 64.14  & 63.16    & 73.54      \\
                          & CRD~\cite{tian2019contrastive}     & 75.11    & 72.21     & 69.73  & 69.11    & 75.65      \\
                          & AT~\cite{zagoruyko2016paying}      & 71.73    & 73.32     & 69.40  & 68.58    & 73.40      \\
                          & KR~\cite{chen2021distilling}       & 77.45    & 77.14     & 70.37  & 69.89    & 77.78      \\ 
\midrule
\multirow{3}{*}{Logits}   & KD~\cite{hinton2015distilling}     & 74.07    & 74.83     & 67.37  & 67.35    & 74.45      \\
                          
                          & DKD~\cite{zhao2022decoupled}       & 76.45    & 76.70     & 69.71  & 70.35    & 77.07      \\ 
                          & NKD$^*$~\cite{sun2024logit}        & 77.11    & 77.12     & 69.98  & 70.45    & 77.37      \\
\midrule
\multicolumn{2}{c|}{Ours}  & \textbf{77.89}    & \textbf{78.10}     &  \textbf{71.02} &  \textbf{72.10}        &\textbf{79.20}       \\
\bottomrule
\end{tabular}}
\end{table*}
This section first provides a brief introduction to our primary experimental settings. 
Subsequently, we evaluate the performance of our method and compare it with previous KD methods. 
Ultimately, we conduct ablation experiments and visual demonstrations to illustrate the efficacy of our method.
\subsection{Experimental Settings}
\paragraph{Datasets.}
(1) CIFAR-100~\cite{krizhevsky2009learning} comprises a well-known image classification dataset, with 60,000  $32 \times 32$ images in total (50,000 for training and 10,000 for validation) from 100 categories;
(2) ImageNet~\cite{deng2009imagenet} is a large-scale classification dataset that consists of 1,000 classes, with nearly 1.3 million training images and 50,000 images for validation;
(3) MS-COCO~\cite{lin2014microsoft}, a challenging object detection dataset, comprises over 200,000 labeled images across 80 categories, including 118,000 training images and 5,000 validation images.
\paragraph{Implementation Details.}
We perform distillation experiments on various neural network architectures, including within homogeneous architecture for teacher-student pairs (e.g., ResNet110 and ResNet32) and heterogeneous architectures (e.g., VGG13 and MobileNet-V2). 
For CIFAR-100, we used 1 NVIDIA 3090 GPU, with a batch size of 64 and a learning rate of 0.05.
For ImageNet, we utilized 4 NVIDIA A100 GPUs, with a batch size of 512 and a learning rate set to 0.2. 
Finally, we also used 4 GPUs for related experiments on MS-COCO, where the batch size was set to 8, and the learning rate was set to 0.01.

\begin{table}[tp]
\centering
\caption{Top-1 and Top-5 accuracy (\%) of student networks on ImageNet validation set.}\label{imagenet}
\scalebox{0.9}{\begin{tabular}{ccc|ccccccc} 
\toprule
\multicolumn{10}{c}{ResNet34 as
  the teacher, ResNet18 as the student}                   \\ 
\midrule
      & Teacher & Student & AT~\cite{zagoruyko2016paying}    & OFD~\cite{heo2019comprehensive}   & CRD~\cite{tian2019contrastive}   & KR~\cite{chen2021distilling}    & KD~\cite{hinton2015distilling}    & DKD~\cite{zhao2022decoupled}     & Ours  \\ 
\midrule
Top-1 & 73.71   & 69.75   & 70.69 & 70.81 & 71.17 & 71.61 & 70.66 & 71.70  & \textbf{72.83}  \\
Top-5 & 91.42   & 89.07   & 90.01 & 89.98 & 90.51 & 90.51 & 89.88 & 90.41  & \textbf{91.38}  \\ 
\midrule
\multicolumn{10}{c}{ResNet50 as the teacher, MobileNetV1
  as the student}                \\ 
\midrule
Top-1 & 76.16   & 68.87   & 69.56 & 71.25 & 72.56 & 72.56 & 68.58 & 72.05  & \textbf{73.25}  \\
Top-5 & 92.86   & 88.76   & 89.33 & 90.34 & 91.00 & 91.00 & 88.98 & 91.05  & \textbf{91.72} \\
\bottomrule
\end{tabular}}
\end{table}
\subsection{Main Results}
\paragraph{Results on CIFAR-100.}
We evaluate our method on CIFAR-100 and compare it with previous methods.
Table \ref{cifar_homo} displays the distillation results between the teacher-student pair with a homogeneous architecture. The results indicate that our method outperforms others in facilitating a more efficient and effective transfer of knowledge from the teacher to the student model. This advantage is evident across all the different pairs, highlighting the robustness and versatility of our approach.

In addition to the experiments on similar architecture pairs, we have also conducted comparative experiments on teacher-student pairs that display significant structural differences. The results of these experiments are presented in Table \ref{cifar_hete}. Similar to the observations made in the homogeneous architecture scenario, we noticed a substantial performance improvement. This result suggests that our method is not only effective in homogeneous architecture scenarios but also in heterogeneous architecture. This further underscores the adaptability and effectiveness of our approach in a wide range of scenarios.

\paragraph{Results on ImageNet.}
In order to showcase the wide-ranging effectiveness of our methodology, we engaged in a series of comparative analyses using a notably intricate dataset, ImageNet.
We report both Top-1 and Top-5 accuracy in Table \ref{imagenet}. We also conducted comparative experiments on homogeneous architectures (ResNet34 and ResNet18) and heterogeneous architectures (ResNet50 and MobileNet). The results in the table show that our method has improved by 0.93\% and 0.24\% compared to the previous SOTA methods on Top-1 accuracy.

\paragraph{Results on MS-COCO.}
We extensively experimented with object detection tasks. Object detection is more complex than classification as it requires not only categorizing objects but also accurately determining their bounding boxes. We employed the widely recognized MS-COCO 2017 dataset for method evaluation for these experiments. Detectron2~\cite{wu2019detectron2}, a popular open-source framework, served as our benchmark. We adopted Faster-RCNN~\cite{ren2015faster} with FPN~\cite{lin2017feature} as the backbone architecture. Performance was assessed on the MS-COCO 2017 validation set using metrics such as $AP$, $AP_{50}$, and $AP_{75}$, with results presented in Table \ref{COCO}. 
The results indicate that our method holds a distinct advantage over approaches utilizing a single type of knowledge. Compared with straightforward hybrids of two knowledge types, UnikD consistently outperforms in most scenarios. However, employing constraints at the distribution level might yield less significant improvements for dense prediction tasks such as object detection than classification tasks. This may be primarily because our method focuses on constraints at the overall distribution level, making such dense prediction tasks less noticeable.
\begin{table*}[tp]
\centering
\caption{Results on MS-COCO. We take Faster-RCNN~\cite{ren2015faster} with FPN~\cite{lin2017feature} as the backbone, and use AP on different settings to evaluate results.}\label{COCO}

\scalebox{0.9}{\begin{tabular}{cc|ccccccccc} 
\toprule
                         &         & \multicolumn{3}{c}{ResNet101 \& ResNet18} & \multicolumn{3}{c}{ResNet101\&  ResNet50} & \multicolumn{3}{c}{ResNet50 \& MobileNet2}  \\ 
\cmidrule{3-11}
                         &         & AP     & AP50   & AP75            & AP     & AP50   & AP75            & AP     & AP50   & AP75          \\ 
\midrule
\multirow{2}{*}{Method}  & Teacher & 42.04~ & 62.48~ & 45.88~          & 42.04~ & 62.48~ & 45.88~          & 40.22~ & 61.02~ & 43.81~        \\
                         & Student & 33.26~ & 53.61~ & 35.26~          & 37.93~ & 58.84~ & 41.05~          & 29.47~ & 48.87~ & 30.90~        \\ 
\midrule
\multirow{3}{*}{Feature} & FitNet~\cite{romero2014fitnets}  
& 34.13~ & 54.16~ & 36.71~          & 38.76~ & 59.62~ & 41.80~          & 30.20~ & 49.80~ & 31.69~        \\
                         & FGFI~\cite{wang2019distilling}
& 35.44~ & 55.51~ & 38.17~          & 39.44~ & 60.27~ & 43.04~          & 31.16~ & 50.68~ & 32.92~        \\
                         & KR~\cite{chen2021distilling}      
& 36.75~ & 56.72~ & 34.00~          & 40.36~ & 60.97~ & 44.08~          & 33.71~ & 53.15~ & 36.13~        \\ 
\midrule
\multirow{3}{*}{Logits}  & KD~\cite{hinton2015distilling}      
& 33.97~ & 54.66~ & 36.62~          & 38.35~ & 59.41~ & 41.71~          & 30.13~ & 50.28~ & 31.35~        \\
                         & DKD~\cite{zhao2022decoupled}     
& 35.05~ & 56.60~ & 37.54~          & 39.25~ & 60.90~ & 42.73~          & 32.34~ & 53.77~ & 34.01~        \\
                         & MLKD~\cite{jin2023multi}   
& 36.03~ & 57.28~ & 38.51~          & 40.15~ & 61.67~ & 44.57~          & 33.83~ & 54.01~ & 35.22~        \\ 
\midrule
\multicolumn{1}{l}{\multirow{2}{*}{Hybrid}} 
& \multicolumn{1}{l|}{FitNet+KD} 
& \multicolumn{1}{c}{34.32}   & \multicolumn{1}{c}{54.52}  & \multicolumn{1}{c}{37.01} 
& \multicolumn{1}{c}{39.03}   & \multicolumn{1}{c}{59.97}  & \multicolumn{1}{c}{42.23} 
& \multicolumn{1}{c}{30.18}   & \multicolumn{1}{c}{50.27}  & \multicolumn{1}{c}{31.59}  \\
\multicolumn{1}{l}{}                        & \multicolumn{1}{l|}{KR+DKD}    
& \multicolumn{1}{c}{37.01}   & \multicolumn{1}{c}{57.53}  & \multicolumn{1}{c}{\textbf{39.85}} 
& \multicolumn{1}{c}{40.65}   & \multicolumn{1}{c}{61.51}  & \multicolumn{1}{c}{44.44} 
& \multicolumn{1}{c}{34.35}   & \multicolumn{1}{c}{54.89}  & \multicolumn{1}{c}{\textbf{36.61}}  \\ 
\midrule
\multicolumn{2}{c|}{Ours}    
&\textbf{37.24}        &\textbf{57.86}        & 39.79       & \textbf{40.76}       & \textbf{61.72}       &   \textbf{44.61}              & \textbf{34.62}       & \textbf{55.14}       & 36.59              \\
\bottomrule
\end{tabular}}
\end{table*}
\subsection{Analyses}
\paragraph{Comparison with Hybrid KD Methods.}
In order to better illustrate the core element of our method, which is the ability to coordinate different types of knowledge at different layers of a network and thus ensure comprehensive and coherent knowledge transfer, we compared our approach with directly hybrid two types of KD methods. We evaluated four different logits-based and feature-based simple hybrid methods and FLG~\cite{hsu2022closer}, which incorporates distillation via logits, features, and gradients but does not unify varying knowledge types.
The results in Table~\ref{homo_hybrid} indicate that combining the two types of distillation methods can significantly enhance the performance of the student network, underscoring the importance of integrating knowledge from different layers. However, compared to the proposed UniKD, which conducts comprehensive and continuous knowledge distillation using a unified type of knowledge, the performance gap remains substantial. By representing the knowledge of each layer of the network in a unified manner, our method can mitigate the inconsistency between different types of knowledge and enhance the distillation performance. This further highlights the necessity of unifying different types of knowledge.
\begin{table*}[tp]
\centering
\caption{Comparing the distillation methods directly combining logits and features, experiments were conducted in six different teacher-student pairs in CIFAR100.}\label{homo_hybrid}
\scalebox{0.9}{\begin{tabular}{cc|cccccc} 
\toprule
\multirow{2}{*}{Model}    & Teacher~ & ResNet56 & ResNet110 & ResNet32$\times$4 & WRN-40-2 & WRN-40-2 & VGG13  \\
                          & Student  & ResNet20 & ResNet32  & ResNet8 $\times$4 & WRN-16-2 & WRN-40-1 & VGG8   \\ 
\midrule
\multirow{2}{*}{Baseline} & Teacher                            & 72.34    & 74.31     & 79.42  & 75.61    & 75.61    & 74.64  \\
                          & Student                            & 69.06    & 71.14     & 72.50  & 73.26    & 71.98    & 70.36  \\ 
\midrule
\multirow{4}{*}{Hybrid}   & KD~\cite{hinton2015distilling}+FitNet~\cite{romero2014fitnets}     
                          & 70.09    & 73.08     & 75.19  & 74.32    & 72.68    & 72.61  \\
                          
                          & KD~\cite{hinton2015distilling} + CRD~\cite{tian2019contrastive}       
                          & 71.63    & 73.75     & 75.46  & 75.64    & 74.38    & 74.29  \\ 
                          & DKD~\cite{zhao2022decoupled} + KR~\cite{chen2021distilling}           
                          & 71.98    &74.23   & 76.45  & 76.19    & 75.52    &75.33\\
                          & MLKD~\cite{jin2023multi} + KR ~\cite{chen2021distilling}
                          & 72.83    &74.52   &78.01   &  \textbf{77.54}   & 76.21    & 75.69 \\
                          & FLG~~\cite{hsu2022closer}           
                          & 71.55    &73.48   & 76.66  & 75.74    & 74.29    &74.28\\
\midrule
\multicolumn{2}{c|}{Ours}                               
                    & {\textbf{73.01}}    & {\textbf{75.36}}     & {\textbf{78.18}}   & {77.49}    & {\textbf{76.35}}    & {\textbf{75.76}}  \\
\bottomrule
\end{tabular}}
\end{table*}
\begin{figure}[t]
\centering
\includegraphics[width=0.9\columnwidth]{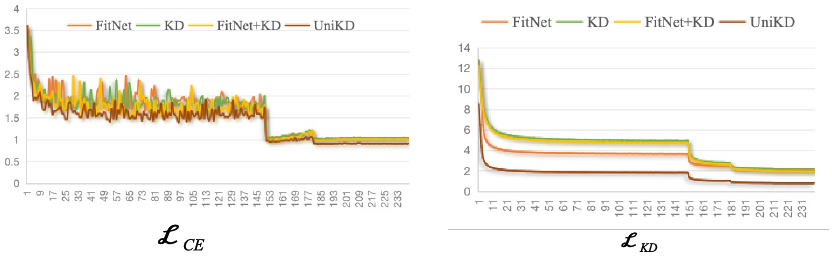}
\caption{A comparative analysis of the convergence of test task loss $\mathcal{L}_{CE}$ and knowledge distillation loss $\mathcal{L}_{KD}$ across various types of distillation methods. All experiments were conducted on the CIFAR100 dataset utilizing a ResNet32$\times$4-ResNet8$\times$4 teacher-student pair architecture.}
\label{loss_vis}
% \vspace{-0.5cm}
\end{figure}
\paragraph{Convergence Analysis.}
Figure~\ref{loss_vis} illustrates the convergence behaviors of different distillation methods. The left side shows the variation of the task loss $\mathcal{L}_{CE}$. It is observable that a single type of method is relatively stable, yet the outcome is sub-optimal. Although a direct hybrid of the two methods results in a loss reduction, the training process exhibits instability, particularly in the early layers of network training. In contrast, our approach ensures stable training and the most favorable results. Additionally, the right side illustrates the variations in distillation loss during the training process.
It can be observed that the proposed UniKD allows the student network to more closely approximate the teacher network, thereby learning more of the teacher’s knowledge.

\begin{table}[tp]
\caption{Ablation study for different modules in UniKD.}\label{main_ablation}
% \vspace{-5pt}
\centering
\begin{tabular}{c|ccc|ccc} 
\toprule
Module & \multicolumn{3}{c|}{ResNet110 \& ResNet32}                                               & \multicolumn{3}{c}{WRN-40-2 \& SN-V1}                                \\
FDP    & \ding{55}     &  \ding{51}     & \ding{51}                             
&\ding{55}             &  \ding{51}     & \ding{51}                   \\
AFF    & \ding{55}     &  \ding{55}     & \ding{51}                   
     & \ding{55}      & \ding{55}      & \ding{51}                   \\ 
\midrule
Acc.   & \multicolumn{1}{c}{73.08~} & \multicolumn{1}{c}{74.42~} & \multicolumn{1}{c|}{\textbf{75.36}~} & \multicolumn{1}{c}{75.21~} & \multicolumn{1}{c}{77.10~} & \multicolumn{1}{c}{\textbf{78.10}~}  \\
\bottomrule
\end{tabular}
% \vspace{-0.1 cm}
\end{table}
\paragraph{Ablation Study.}
To validate the effectiveness of each module in our method, we analyzed their contributions individually. Table~\ref{main_ablation} shows distillation results for both homogeneous (i.e., ResNet110 and ResNet32) and heterogeneous architectures (i.e., WRN-40-2 and ShuffleNet-V1), underlining our method's efficacy. The results confirm that our carefully crafted modules significantly enhance the student model's performance.
In our initial approach, we apply the Feature Distribution Processor (FDP) to the output features of the last intermediate layer to unify different types of knowledge, significantly enhancing the performance of the student model. The results indicate that our method achieves substantial improvements. However, due to the lack of integration of information from different layers, the utilization of the teacher's knowledge still needs to be improved. Therefore, we incorporate the Adaptive Feature Fusion (AFF) module to integrate intermediate features from different layers, achieving a comprehensive and unified knowledge distillation. The results demonstrate the effectiveness of the AFF module.

\paragraph{Comparison with Teacher Model.}
The results in Table~\ref{compare_teacher} show that while most methods improve upon the original student model, they often do not outperform the teacher model. However, with our UniKD, the distilled student model not only significantly outperforms its original version but also exceeds the teacher model in most cases, achieving an average improvement of 0.57\% across six teacher-student pairs. The results highlight UniKD's efficacy in capturing essential knowledge from the teacher network and minimizing redundant information transfer, leading to a lightweight student model that surpasses the more cumbersome teacher model.
\begin{figure}[tp]
\centering
\includegraphics[width=0.7\columnwidth]{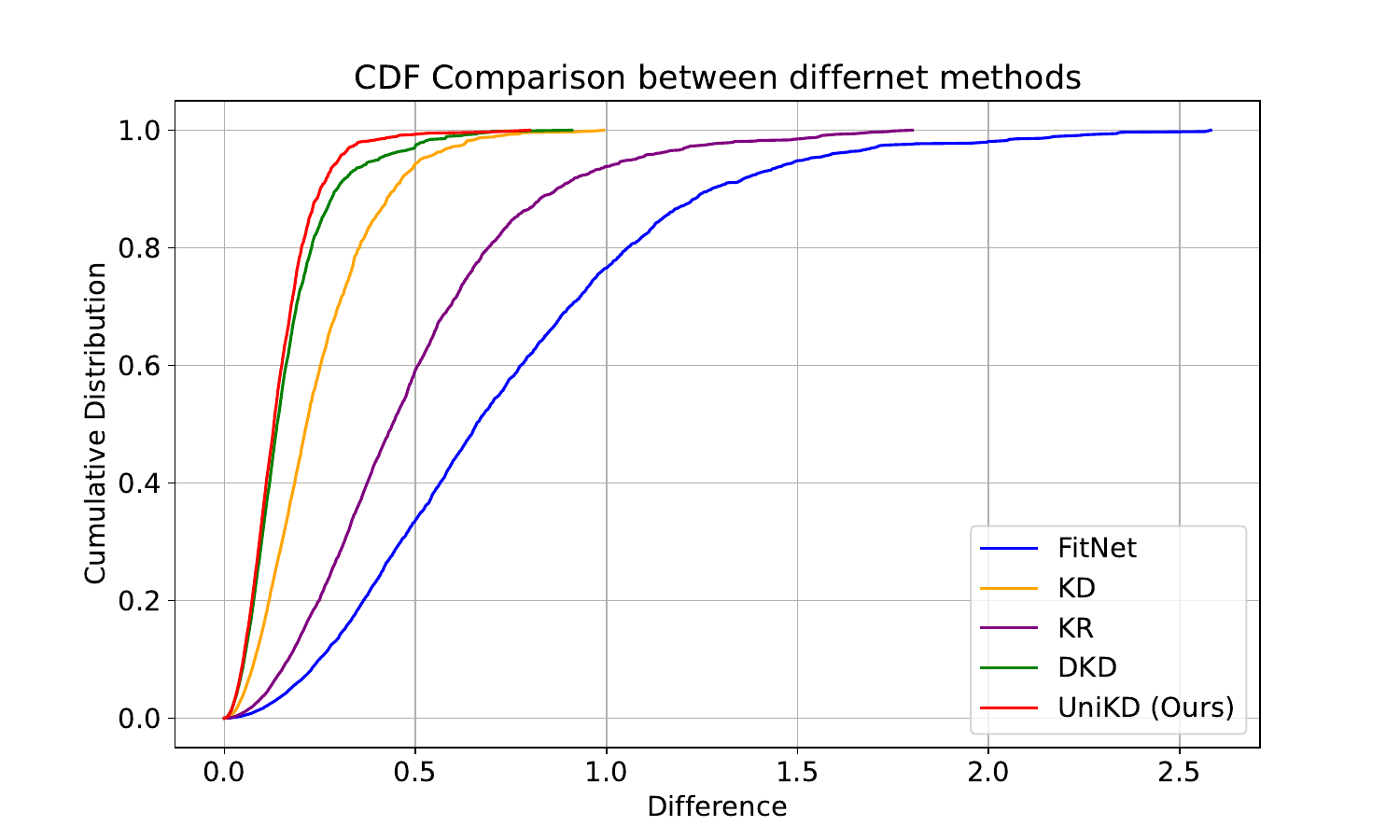}
\caption{CDF curve of the differences in logits output by teacher and student in correspondence with various methods.}
\label{cdf}
% \vspace{-0.3cm}
\end{figure}
\begin{table}[tp]
\centering
\caption{
% Performance gap among the teacher model, the original student model, and the distilled student model. 
We calculated the average metrics the deviations from both the student and the teacher of different teacher-student pairs in Table~\ref{cifar_homo}.
% % We recorded the gap between the original student model and the teacher model for different distillation methods on six teacher-pairs.
% Note that when the distilled student model outperforms the original model, the gap is positive. 
}\label{compare_teacher}
% \begin{tabular}{cc|c|cc} 
% \toprule
%                          &         & Avg   & $\Delta_{stu}$ & $\Delta_{tea}$   \\ 
% \midrule
% \multirow{2}{*}{Method}  & Teacher &\multicolumn{3}{c}{75.32}        \\
%                          & Student &\multicolumn{3}{c}{71.38}       \\ 
% \midrule
% \multirow{4}{*}{Feature} & FitNet~\cite{romero2014fitnets}      & 71.77 & 0.39 & -3.55  \\
%                          & CRD~\cite{tian2019contrastive}       & 71.73 & 0.35 & -3.59  \\
%                          & AT~\cite{zagoruyko2016paying}        & 72.43 & 1.05 & -2.89  \\
%                          & KR~\cite{chen2021distilling}         & 74.58 & 3.20  & -0.74  \\ 
% \midrule
% \multirow{3}{*}{Logits}  & KD~\cite{hinton2015distilling}       & 73.09 & 1.71 & -2.23  \\
%                          & DKD~\cite{zhao2022decoupled}         & 74.69 & 3.31 & -0.63  \\
%                          & MLKD~\cite{jin2023multi}             & 75.09 & 3.71 & -0.23  \\ 
% \midrule
% \multicolumn{2}{c|}{Ours}       & \textbf{75.89} & \textbf{+4.51} & \textbf{+0.57}   \\
% \bottomrule
% \end{tabular}
\begin{tabular}{c|cccc|ccc|c|c} 
\toprule
\multicolumn{1}{l|}{} & \multicolumn{4}{c|}{Feature}   & \multicolumn{3}{c|}{Logits} & \multicolumn{1}{c|}{Hybrid} & Ours   \\ 
\midrule
                      & FitNet & CRD   & AT    & KR    & KD    & DKD   & NKD        & DKD+KR             & UniKD  \\ 
\midrule
Avg                   & 71.77  & 71.73 & 72.43 & 74.58 & 73.09 & 74.69 & 75.09       & 74.95                 & \textbf{75.89}  \\ 
\midrule
$\Delta_{stu}$                   & 0.39   & 0.35  & 1.05  & 3.2   & 1.71  & 3.31  & 3.40        & 3.57                 & \textbf{4.31}   \\
$\Delta_{tea}$                   & -3.55  & -3.59 & -2.89 & -0.74 & -2.23 & -0.63 & -0.37       & -0.37                 & \textbf{0.54}   \\
\bottomrule
\end{tabular}
% \vspace{-0.5 cm}
\end{table}
\paragraph{Visualizations.}
We analyzed the Cumulative Distribution Function (CDF) curves of different methods, where the $x$-axis indicates the difference magnitude between teacher and student model outputs, and the y-axis shows the proportion of differences not exceeding each $x$-axis value. As Figure~\ref{cdf} demonstrates, our method's CDF curve swiftly rises to 1, which suggests a high degree of similarity between the outputs of the student and teacher models, implying effective learning and enhanced performance of the student model.

\begin{figure}[ht]
\centering
\includegraphics[width=0.9\columnwidth]{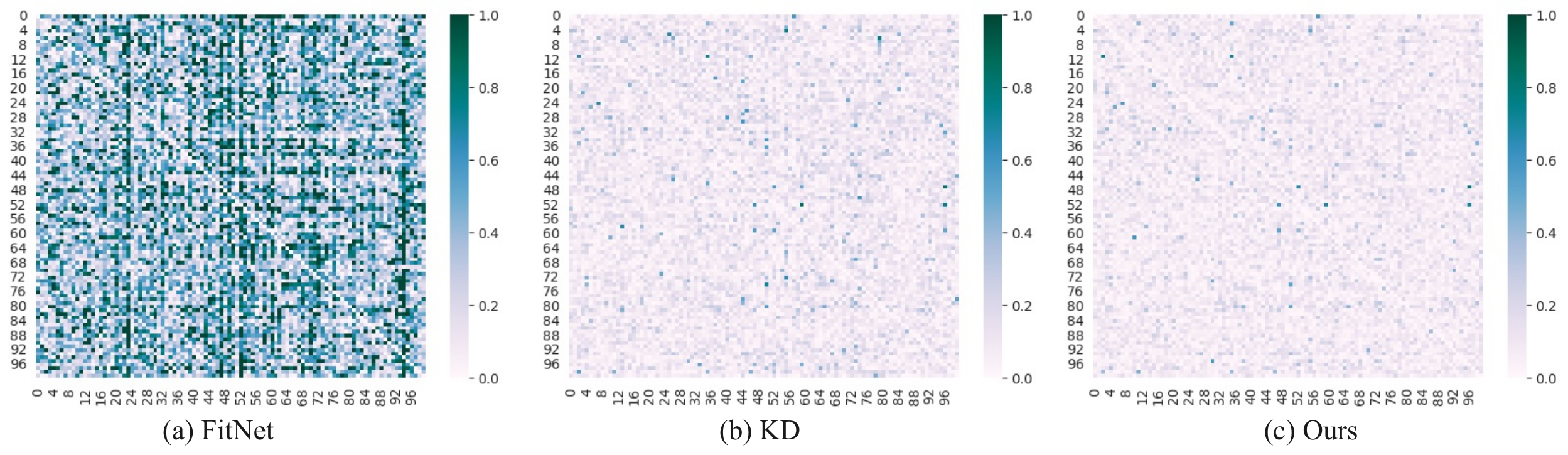}
\caption{Difference of correlation matrices of student and teacher logits.}
\label{featvis}
% \vspace{-1cm}
\end{figure}
We also visualize the differences in the correlation matrices between student and teacher logits, as illustrated in Figure~\ref{featvis}. UniKD enables the student to produce logits that are more similar to those of the teacher compared to FitNet and KD, thereby achieving superior distillation performance.
\section{Conclusion}
\label{conclusion}
We analyzed that different network layers contain different types of knowledge, thus necessitating knowledge distillation at various layers. However, using different types of constraint methods can lead to unclear optimization objectives. Therefore, it is essential to unify the different types of knowledge.
Based on this, We present a novel KD framework harmonizing various knowledge sources. It adapts intermediate knowledge to align with final-layer logits, providing a uniform knowledge form and enabling efficient KD throughout the network stages. We tested it extensively with various teacher-student pairs on different tasks. Our method outdid both logits and feature-based KD. It successfully acquired sufficient knowledge from the teacher network in various analytical experiments.

\section*{Acknowledgements}
This work was supported by the National Key R \& D Program of China (2022ZD0161800),
and the National Natural Science Foundation of China (62271203).

% ---- Bibliography ----
%
% BibTeX users should specify bibliography style 'splncs04'.
% References will then be sorted and formatted in the correct style.
%
\bibliographystyle{splncs04}
\bibliography{main}
\end{document}